\documentclass[lettersize,journal]{IEEEtran}
\usepackage{amsmath,amsfonts}

\usepackage{array}
\usepackage{subfigure}
\usepackage{textcomp}
\usepackage{stfloats}
\usepackage{url}
\usepackage{verbatim}
\usepackage{graphicx}
\def\BibTeX{{\rm B\kern-.05em{\sc i\kern-.025em b}\kern-.08em
    T\kern-.1667em\lower.7ex\hbox{E}\kern-.125emX}}
\usepackage{balance}

\usepackage[numbers]{natbib}
\usepackage{enumitem}
\usepackage{multirow}
\usepackage{bm}
\usepackage{bbding}
\usepackage{pifont}
\usepackage{makecell}
\usepackage{algorithm}  
\usepackage{algpseudocode} 
\begin{document}
\title{Quiz-based Knowledge Tracing}
\author{Shuanghong Shen, Enhong Chen,~\IEEEmembership{Senior Member,~IEEE}, Bihan Xu, Qi Liu, Zhenya Huang, Linbo Zhu, Yu Su
	
	\IEEEcompsocitemizethanks{
		\IEEEcompsocthanksitem S.~Shen, E.~Chen~(corresponding author), B.~Xu, and Z.~Huang are with the Anhui Province Key Laboratory of Big Data Analysis and Application, School of Data Science \& School of Computer Science and Techonology, University of Science and Technology of China \& State Key Laboratory of Cognitive Intelligence, Hefei, Anhui, 230026, China.
		\protect Email: \{closer, xbh0720\}@mail.ustc.edu.cn, \{cheneh, huangzhy\}@ustc.edu.cn
	}
	
	\IEEEcompsocitemizethanks{
		\IEEEcompsocthanksitem Q.~Liu is with the Anhui Province Key Laboratory of Big Data Analysis and Application, School of Data Science \& School of Computer Science and Techonology, University of Science and Technology of China \& State Key Laboratory of Cognitive Intelligence \& Institute of Artificial Intelligence, Hefei Comprehensive National Science Center, Hefei, Anhui, 230026, China.
		\protect Email: qiliuql@ustc.edu.cn
	}
	
	\IEEEcompsocitemizethanks{
		\IEEEcompsocthanksitem Y.~Su is with the School of Computer Science and Technology, Hefei Normal University \& Institute of Artificial Intelligence, Hefei Comprehensive National Science Center, Hefei, Anhui, 230601, China.
		\protect Email: yusu@hfnu.edu.cn
	}
	\IEEEcompsocitemizethanks{
		\IEEEcompsocthanksitem L.~Zhu is with the Institute of Artificial Intelligence, Hefei Comprehensive National Science Center, Hefei, Anhui, 230026, China.
		\protect Email: lbzhu@iai.ustc.edu.cn
	}
	
}
\markboth{}%
{How to Use the IEEEtran \LaTeX \ Templates}

\maketitle

\begin{abstract}
		Knowledge tracing (KT) aims to assess individuals' evolving knowledge states according to their learning interactions with different exercises in online learning systems (OIS), which is critical in supporting decision-making for subsequent intelligent services, such as personalized learning source recommendation. Existing researchers have broadly studied KT and developed many effective methods. However, most of them assume that students' historical interactions are uniformly distributed in a continuous sequence, ignoring the fact that actual interaction sequences are organized based on a series of quizzes with clear boundaries, where interactions within a quiz are consecutively completed, but interactions across different quizzes are discrete and may be spaced over days. In this paper, we present the Quiz-based Knowledge Tracing (QKT) model to monitor students' knowledge states according to their quiz-based learning interactions. Specifically, as students' interactions within a quiz are continuous and have the same or similar knowledge concepts, we design the adjacent gate followed by a global average pooling layer to capture the intra-quiz short-term knowledge influence. Then, as various quizzes tend to focus on different knowledge concepts, we respectively measure the inter-quiz knowledge substitution by the gated recurrent unit and the inter-quiz knowledge complementarity by the self-attentive encoder with a novel recency-aware attention mechanism. Finally, we integrate the inter-quiz long-term knowledge substitution and complementarity across different quizzes to output students' evolving knowledge states. Extensive experimental results on three public real-world datasets demonstrate that QKT achieves state-of-the-art performance compared to existing methods. Further analyses confirm that QKT is promising in designing more effective quizzes.
\end{abstract}


\section{Introduction}
\IEEEPARstart{O}{nline} learning systems (OIS) have been playing an increasingly important role in satisfying individuals' growing demands for intelligent educational services \citep{karampiperis2004adaptive, bates2007self}, e.g., personalized learning source recommendation  \citep{tarus2018knowledge, liu2019exploiting}. Knowledge tracing (KT), which aims to monitor students’ dynamic knowledge states in learning based on their learning interactions on OIS, is one of the fundamental research tasks to provide guidance for these intelligent services \citep{corbett1994knowledge}. In recent years, an increasing amount of attention has been abstracted to this emerging research area \citep{liu2021survey}. 

Generally, OIS assigns exercises related to different Knowledge Concepts (KCs, e.g., \textit{Adding and Subtracting Fractions}) for students to answer so that they can acquire the required knowledge. According to students' interactions, i.e., their performance on different exercises, researchers have designed different KT methods to infer their knowledge states and predict their future performance. Subsequently, we can enhance the learning and teaching efficiency by adopting targeted teaching strategies for each student in accordance with their knowledge states. 
In the literature, most of existing methods measure students' knowledge states through sequence modeling. For example, Bayesian knowledge tracing (BKT) formalized the learning process as the Markov process and utilized the Hidden Markov Model to assess the dynamic knowledge state \citep{corbett1994knowledge}. Deep knowledge tracing (DKT) further introduced RNNs/LSTMs \citep{hochreiter1997long} to conduct sequence modeling on students' learning interactions \citep{piech2015deep}. Many subsequent studies have improved BKT and DKT in different aspects, such as considering students' individual characteristics \citep{pardos2010modeling, yudelson2013individualized, shen2020convolutional}, utilizing more side information \citep{wang2012leveraging,gonzalez2014general}, incorporating the structure of KCs \citep{nakagawa2019graph, tong2020structure}. Moreover, some latest works presented new architectures to solve the KT problem, such as using memory networks to store and update the knowledge state \citep{zhang2017dynamic}, applying the attention mechanism to capture the knowledge dependency of learning interactions \citep{pandey2019self, CAKT}.  

\begin{figure*}

	\centering
	\includegraphics[width= 0.92\textwidth]{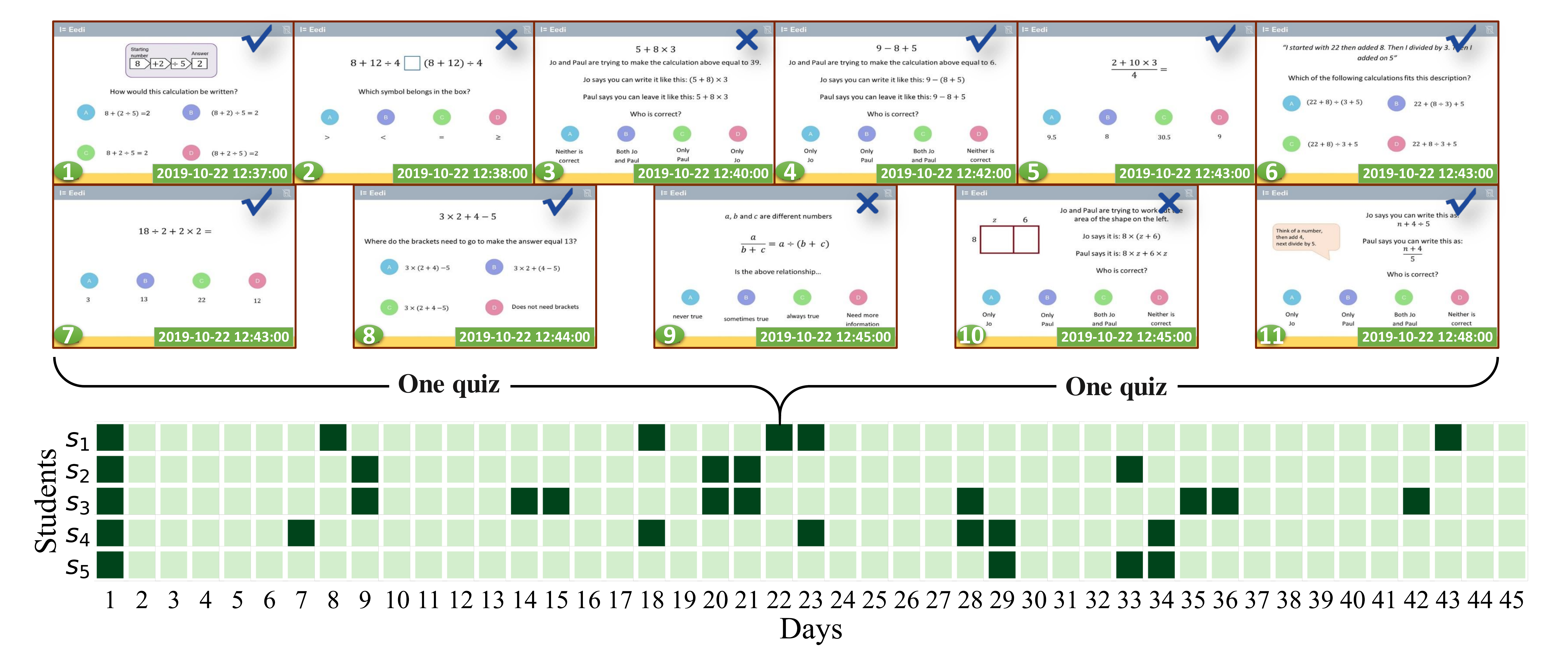}
	\vspace{-0.3cm}
	\caption{Real examples for students' interaction sequences from the perspective of quizzes. In the lower block diagram, we recorded students' interactions based on days and the first day was the starting point of the sequence. Here, a square denotes for one day, where dark squares denote that students finished one or more quizzes in the corresponding day and light squares represent they did not answer any exercises. The lower part indicates that students' interaction sequences are organized based on a series of quizzes with clear boundaries. In the upper part, we give the visualization of $s_1$'s continuous interactions on a specific quiz of 11 different exercises related to the same KC: \textit{Basic Arithmetic}. $s_1$ spent about 11 minutes completing this quiz.}
	\label{case1}
	\vspace{-0.5cm}
\end{figure*}

However, most of existing KT methods, including the above mentioned ones, assume that students' historical interactions are uniformly distributed in a continuous sequence, which does not conform to the reality. Actually, exercises in OIS are assigned to students in the form of quizzes rather than individually \citep{juhavnak2019using, jensen2021you}. Specifically, the quiz is defined as an informal test of specific knowledge, which is consisted of a number of exercises of the same or similar KCs \citep{spanjers2015promised}.  Therefore, students' historical interactions are only continuous within the same quiz, and there are clear boundaries between different quizzes, which may be spaced over several days. 
To better illustrate, we give some real examples of students' interaction sequences from the perspective of quizzes in Figure \ref{case1}. These examples are extracted from the real learning data in Eedi \citep{wang2020diagnostic}, an OIS that millions of students interact with daily around the globe. In Figure \ref{case1}, we recorded students' interactions based on days and the first day denoted for the starting point of the interaction sequence. The lower part of Figure \ref{case1} clearly indicates that students' historical interactions are quiz-based and discrete. Besides, in the upper part of Figure \ref{case1}, we give the details of $s_1$'s interactions that occurred continuously within a specific quiz: the whole quiz answering process took about 11 minutes and all of the exercises in this quiz have the same KC: \textit{Basic Arithmetic}. 
In summary, students' learning interactions within the same quiz are continuous over a short period of time, while these across different quizzes are discrete with certain intervals. 

In this paper, we argue that it is critical and beneficial to consider the quiz-based organization style of students' learning interactions in KT. Unfortunately, there are many technical and domain challenges to be solved along this line.  
First, as we have mentioned above, exercises in the same quiz usually have similar KCs and students' related interactions are continuous over a short period of time, it is a nontrivial problem to capture the intra-quiz short-term knowledge influence.  For example, compared to a hard previous exercise, answering an easy previous exercise should have different influence on students' performance on the present exercise within a quiz \citep{bard2017effect}.
There are more challenges when coming to different quizzes, as they are discrete and usually have various KCs. Specifically, if a recent quiz has similar KCs to previous quizzes, students' interactions on previous similar quizzes may become unreliable and be replaced by the recent ones, how to capture such inter-quiz long-term knowledge substitution? Besides, if a recent quiz has new KCs that have never appeared in previous quizzes, how to integrate these quizzes related to various KCs, i.e., to measure the inter-quiz long-term knowledge complementarity?

To achieve our primary goal of realizing quiz-based KT with addressing the above challenges, we propose the Quiz-based Knowledge Tracing (QKT) model in this paper, which measures students' knowledge states by exploring their quiz-based learning interactions. Specifically, we first design the adjacent gate to control the knowledge influence between adjacent interactions within the same quiz. Considering that students' average performance on a quiz reflects their knowledge states on the quiz-related KC, we further perform the global average pooling operation for each quiz. For example, the student $s_1$ in Figure \ref{case1} got 7 correct answers and 4 wrong answers on the contained 11 exercises related to the same KC: \textit{Basic Arithmetic},  $s_1$'s knowledge state with respect to \textit{Basic Arithmetic} should be approximately $\frac{7}{11}$. 
Then, we directly utilize the Gated Recurrent Units (GRU) \citep{cho2014learning} to assess the inter-quiz long-term knowledge substitution, which memorizes students' interactions on recent quizzes and forgets the remote ones. Besides, we present the self-attentive encoder to measure the inter-quiz knowledge complementarity, which preserves and fuses students' interactions on all historical quizzes. To measure students' varying degrees of knowledge loss on historical quizzes, we design a novel recency-aware attention mechanism in the self-attentive encoder. Finally, we can get students' evolving knowledge states by combining the inter-quiz long-term knowledge substitution and complementarity between different quizzes.

Our main contributions are summarized as follows:
\begin{itemize}[leftmargin=*,itemsep=3pt]
	\item{} We firstly focus on the quiz-based organization style of students' learning interactions in OIS for the KT task. We summarize the feature of the quiz-based interaction sequence, i.e., it is continuous over a short period of time within a quiz and discrete with certain intervals across different quizzes.  
	We further give detailed analysis of three public real-world datasets collected from different OIS from the perspective of quizzes in Section \ref{datasets}. 
	\item{} We propose a novel Quiz-based Knowledge Tracing model to assess students' dynamic knowledge states by exploring their quiz-based interaction sequences. In QKT, we respectively measure the intra-quiz short-term knowledge influence and inter-quiz long-term knowledge substitution and complementarity.
	\item{} We conduct extensive experiments to verify the effectiveness of QKT, the results indicate that QKT has superior performance to existing methods. Further analyses indicate that QKT can be utilized to help design more effective quizzes.
\end{itemize}
\section{Related Works}
In this section, we introduce existing related works from two categories: knowledge tracing, and cognitive diagnosis.

\subsection{Knowledge Tracing}
With the development of OIS, the significance of monitoring students' knowledge states is becoming increasingly prominent \citep{liu2021survey}, which was first formalized as the knowledge tracing task by \citet{corbett1994knowledge}. They proposed the BKT model, assuming the learning process as a Markov process and using students' observed interaction sequences to infer their latent knowledge states. Then, researchers enriched and developed BKT in many aspects. 
For example, individualizing the parameter in BKT for each student \citep{pardos2010modeling, yudelson2013individualized}, considering the tutor intervention of OIS \citep{lin2016intervention}, and incorporating students' forgetting effect \citep{nedungadi2015incorporating}.
In recent years, the advances of deep learning (DL) have boosted the neural network based KT models. Specifically, DKT introduced RNNs/LSTMs to model the students' knowledge states in a sequence manner \citep{piech2015deep}. Then, DKVMN used memory networks to store and update students' latent knowledge states on specific KCs \citep{zhang2017dynamic}. Some researchers considered the natural structure within the KCs, and proposed to use GNNs to capture the influence integration of the knowledge state between different KCs \citep{wu2019comprehensive, nakagawa2019graph, tong2020structure}. Besides, some studies noticed that students' historical related interactions had more impacts on their future performance. Therefore, they introduced the attention mechanism to model the knowledge dependencies in learning. For example, \citet{pandey2019self} applied the Transformer \citep{vaswani2017attention} to trace students' knowledge states, \citet{CAKT} incorporated self-attention mechanism with monotonic assumption. \citet{mfdakt} applied the dual-attentional mechanism to model students’ learning progress based on multiple factors. There were also many works focused on the representation of exercises. 
For example, learning the semantic representations of exercises from their text contents \citep{huang2019ekt, RKT}. \citet{PEBG} turned to pursue pre-trained exercise embeddings from exercise-KC relations, the exercise similarity, KC similarity, and the exercise difficulties together. \citet{shen2022assessing} modeled the exercise difficulty effect and designed an adaptive sequential neural network to match the exercise difficulty with the knowledge state. Recently, researchers further explored students' learning process. \citet{hawkesKT} presented the Hawkes process to adaptively model temporal cross-effects in learning. \citet{shen2021learning} proposed to model students' learning gains and forgetting in learning for calculating their dynamic knowledge states. \citet{long2021tracing} estimated students' individual cognition level and knowledge acquisition in learning.

In summary, most of existing KT methods follow the paradigm of sequence modeling. They assumed that students' historical interaction sequences are uniformly distributed in a continuous sequence, which neglects the fact that students' interactions are quiz-based with clear boundaries. Therefore, interactions across different quizzes are actually discrete. Although some works have noted the significance of the interaction's timestamp \citep{shin2021saint+, shen2021learning}, they were limited to simply utilizing the time information as additional features. \citet{ke2022hitskt} further split students' historical interactions into sessions based on fixed time duration and performed session-aware KT. However, it was also inconsistent with the quiz-based organization of students' interactions and damaged the knowledge correlation intra- and inter-quiz.

\subsection{Cognitive Diagnosis}
Cognitive diagnosis (CD) is also concerned with assessing individuals’ knowledge states based on their behaviors \cite{Liu_2021}. In contrast to KT, CD is often applied in testing scenarios, which utilizes all historical interactions to learn each student’s static knowledge state. 
Specifically, the item response theory (IRT) is one of the classical CD models \cite{hambleton1991fundamentals}, which used a logistic regression model to estimate students' knowledge states:
\begin{equation}
	p = c +  \frac {1 - c}{1 + e^{-(\theta - \beta)}},
\end{equation}
where $c$ is the random guessing probability, $\theta$ is the knowledge state, $\beta$ is the exercise difficulty. However, $\theta$ in IRT is a single value, which cannot reflect students' knowledge states on various KCs. Therefore, multidimensional item response theory (MIRT) was proposed to use a multidimensional vector to represent students' knowledge states on different KCs \cite{reckase1997past, Reckase_2006}.
In recent years, deep learning has been widely employed for cognitive diagnosis \cite{Cheng_2019, 2019Deep, Wang2020neural}.  For example, the Neural Cognitive Diagnosis (NCD) attempted to utilize neural networks to model the student-exercise interactions. NCD's general framework can be formulated as:
\begin{equation}
	p = \phi_n(...\phi_1(F^s, F^{kc}, F^{other}, \theta_f)),
\end{equation}
where $\phi$ denotes the neural network used to model the student-exercise interactions. $F^s$ is the knowledge state, $F^{kc}$ means the KC factor, $F^{other}$ denotes other factors (such as exercise difficulty), $\theta_f$ denotes all learnable parameters. 
Subsequently, researchers have made extensions to NCD in different aspects, such as incorporating students' abundant context information \citep{Zhou_2021} and measuring the hierarchical relations among students, exercises, and KCs \citep{Gao_2021}.
However, CD has an underlying assumption that all of students' interactions are equally important to their knowledge states. This assumption is reasonable within a single test/quiz, but may not be reliable for students' quiz-based interaction sequences in reality.

%\subsection{Session-based Recommendations}

\begin{figure*}
	
	\centering{\includegraphics[width=\textwidth]{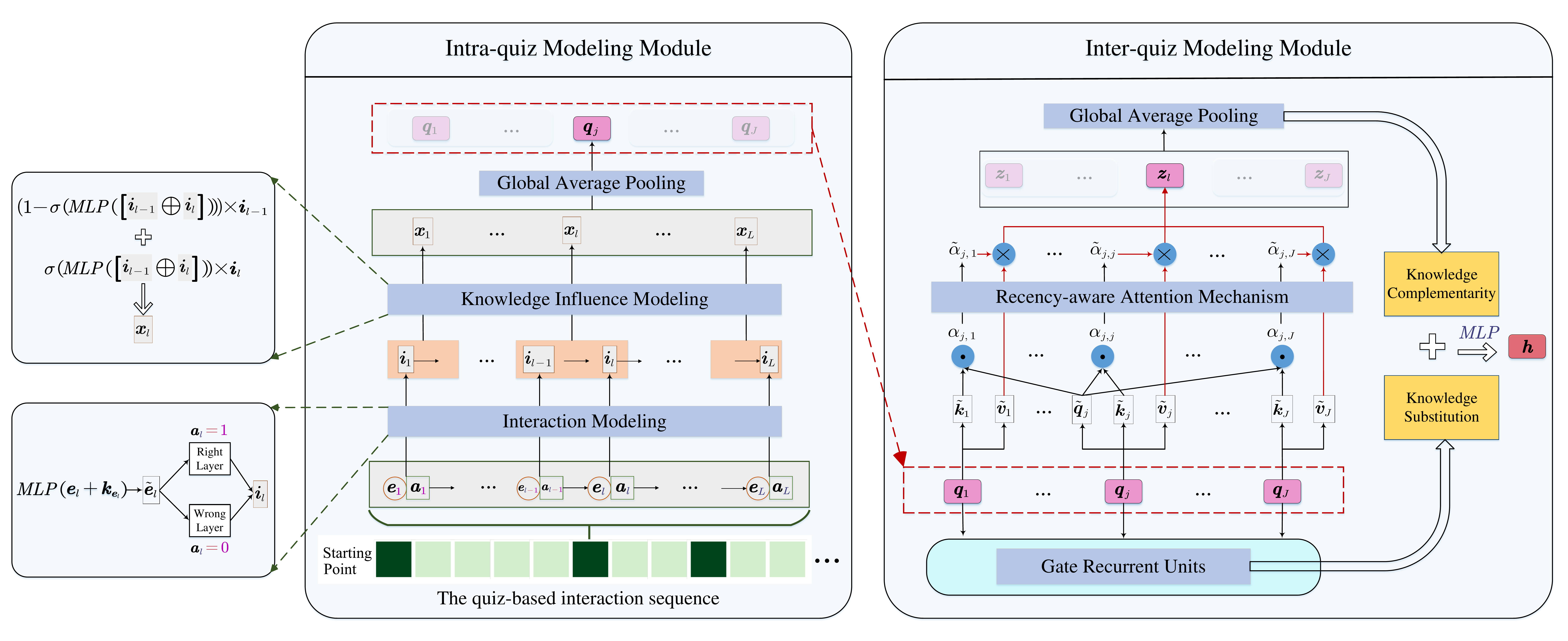}}
	\vspace{-0.5cm}
	\caption{The architecture of the QKT model.}
	\label{model}
	\vspace{-0.5cm}
\end{figure*}

\section{Problem Statement}
The OIS contains multiple basic elements, including \textit{students}, \textit{exercises}, \textit{KCs}, and \textit{student-exercise interactions}. Supposing that all the students together in a dataset form the student set, namely $\mathbb{S} = \{s_1, s_2, ..., s_S\}$, all the exercises together form the exercise set, namely $\mathbb{E} = \{e_1, e_2, ..., e_E\}$, all the KCs together form the KC set, namely $\mathbb{K} = \{k_1, k_2, ..., k_K\}$,  where $s, e, k$ respectively denote a student, an exercise, and a KC, $S, E, K$ respectively denote the number of students, exercises, and KCs. In general, each exercise is related to specific KCs and we use the Q-matrix given by educational experts to indicate the exercise-KC relations. The Q-matrix is made up of ones and zeros, where one means the corresponding exercise and KC are related, otherwise it is zero. For a specific student $s$, the \textit{student-exercise interaction} is the most basic unit. We have $i$ denotes an interaction of $s$, which includes an exercise $e$ and the answer $a$ given by $s$ on $e$, i.e., $i = (e,a|s)$. Here the answer $a$ is a binary correctness label (1 represents correct and 0 means incorrect). The quiz is the basic organization form of exercises in OIS. 
Generally, a quiz is presented as an informal test of specific knowledge, which is made up of multiple exercises with the same or similar KCs. For example, the given quiz in Figure \ref{case1} contains 11 exercises with the same KC: \textit{Basic Arithmetic}. Noting that the quiz may have different names in different systems, e.g., it is called the \textit{assignment} in ASSISTments \citep{feng2009addressing} and CodeWorkout \citep{edwards2017codeworkout}. For the sake of convenience, we uniformly use the name quiz throughout this paper. We have $q$ denotes the student's interactions on a quiz, i.e., $q = \{i_1, i_2, ..., i_L\}$, $L$ is the length of the quiz $q$ (i.e., the length of interactions in $q$) and the subscripts from $1$ to $L$ represent the order of each interaction.  Let $U$ denote the student's whole interaction sequence, we can represent $U$ as the quiz set, i.e., $U = \{q_1, q_2, ..., q_J\}$, $J$ is the number of quizzes and the subscripts from $1$ to $J$ represent the order of each quiz. Then, we can formally formulate the task of quiz-based knowledge tracing as follows:

\paragraph{\textbf{Problem Formalization.}}  Given a student's sequential interactions $U = \{q_1, q_2, ..., q_J\}$ on multiple quizzes, the quiz-based knowledge tracing task aims to assess the student's dynamic knowledge states across different quizzes and predict her performance on new exercises in future quizzes.

\section{Quiz-based Knowledge Tracing}
In this section, we present the proposed QKT model in detail and indicate how to measure the intra-quiz short-term knowledge influence, as well as the inter-quiz long-term knowledge substitution and complementarity. 
The architecture of QKT is depicted in Figure \ref{model} and  Algorithm \ref{a1}, which is mainly consisted of intra-quiz modeling module and inter-quiz modeling module. Specifically, in the intra-quiz modeling module, we mainly focus on exploring the knowledge influence between students' adjacent interactions within a quiz and capturing their overall knowledge states for each quiz. Then, in the inter-quiz modeling module, we turn to measure the knowledge substitution and knowledge complementarity between different quizzes, which will be finally integrated together to output students' evolving knowledge states across different quizzes. 

\subsection{The Intra-quiz Modeling Module}
In this module, given the student's interactions on the quiz $q = \{i_1, i_2, ..., i_L\}$, we aim to measure the intra-quiz short-term knowledge influence and output the quiz vector $\bm{q}$ that represents the student's knowledge states for the quiz $q$.

\subsubsection{Interaction Modeling}
A quiz usually contains multiple interactions, which are consisted of \textit{exercise-answer} pairs. Therefore, we first conduct interaction modeling in this part.

Specifically, we use the embedding matrix $\bm{E} \in \mathbb{R}^{E \times d_e}$ to represent all exercises in a dataset. Therefore, for the specific exercise $e_l$ in the quiz $q$, we can obtain its embedding $\bm{e}_l$ from $\bm{E}$. Besides, considering that exercises have different KCs, we also use an embedding matrix $\bm{K} \in \mathbb{R}^{K \times d_k}$ (here we set $d_e$ to be equal to $d_k$) to represent all KCs, so that we can also obtain the KC embedding $\bm{k}_{e_l}$ of $e_l$ from $\bm{K}$. Then, we combine the exercise embedding $\bm{e}_l$ and its KC embedding $\bm{k}_{e_l}$ by a multi-layer perceptron (MLP) to represent the complete exercise:
\begin{equation}  \label{eq1}
	\bm{\tilde{e}}_l = \text{Relu}(\bm{W}_1^T (\bm{e}_l + \bm{k}_{e_l})+ \bm{b}_1), \\
\end{equation}
where we have $\bm{\tilde{e}}_l \in \mathbb{R}^{d_e}$ denotes the complete exercise vector, $\bm{W}_1 \in  \mathbb{R}^{d_e \times d_e}$ and $\bm{b}_1 \in \mathbb{R}^{d_e}$ are trainable parameters.

Subsequently, considering the binary answer value, we respectively present the right layer and the wrong layer to distinguish the different effects of two binary answers. Then, the \textit{exercise-answer} pair can be represented as follows:
\begin{equation}  \label{eq2}
	\bm{i}_l = 
	\begin{cases}
		\bm{W}_{r}^T\bm{\tilde{e}}_l + \bm{b}_r, & \text{if} \quad   a_l = 1, \\
		\bm{W}_{w}^T\bm{\tilde{e}}_l+ \bm{b}_w, & \text{if} \quad   a_l = 0, \\
	\end{cases}
\end{equation}
where we have $\bm{i}_l \in \mathbb{R}^{d_i}$ denotes the interaction vector.  $\bm{W}_r \in  \mathbb{R}^{d_e \times d_i}$ and $\bm{b}_r \in \mathbb{R}^{d_i}$ are trainable parameters for the right layer,  $\bm{W}_w \in  \mathbb{R}^{d_e \times d_i}$ and $\bm{b}_w \in \mathbb{R}^{d_i}$ are trainable parameters for the wrong layer. 

\subsubsection{Knowledge Influence Modeling} After finishing the interaction modeling, we proceed to measure the knowledge influence of interactions within a quiz. Concretely, such knowledge influence mainly exists in adjacent interactions from many aspects. For example, spending different energy on the previous exercise should have a different impact on students' performance on the current exercise \citep{joseph2005engagement}. Besides, previous hard exercises bring more negative effects than easy exercises, i.e., a learning effect occurs when easy exercises come before harder exercises and a fatigue effect occurs when exercises come in a hard-to-easy order \cite{Kingston1984, bard2017effect}. 

In our proposed QKT model, we measure the above adjacent multi-aspect knowledge influence in an uniform manner. Specifically, we design the adjacent gate according to the influence between two adjacent interactions, which is then applied to control how much information should be respectively extracted from the previous and current interactions. The calculating formulas are:
\begin{equation} \label{eq3}
	\begin{aligned}
		\bm{\varGamma}_l &=  \sigma(\bm{W}_2^T (\bm{i}_{l-1} \oplus \bm{i}_l))+ \bm{b}_2, \\
		\bm{x}_l &= \bm{\varGamma}_l \cdot \bm{i}_l + (1 - \bm{\varGamma}_l) \cdot \bm{i}_{l-1}, \\
	\end{aligned}
\end{equation}
where we have $\bm{\varGamma}_l \in \mathbb{R}^{d_i}$ denotes the adjacent gate, $\bm{x}_l \in \mathbb{R}^{d_i}$ denotes the combined vector of two adjacent interactions. $\sigma$ is the \textit{sigmoid} activation function, $\cdot$ is the element-wise product operation,  $\oplus$ means vector concatenation, $\bm{W}_2 \in  \mathbb{R}^{2d_i \times d_i}$ and $\bm{b}_2 \in \mathbb{R}^{d_i}$ are trainable parameters. 

\subsubsection{Global Average Pooling} \label{413}
In this part,  we further explore to making trade-offs between all interactions in a quiz to get the quiz vector that represents the student's overall knowledge states for the quiz $q$. As interactions in a quiz have the same or similar KCs, each interaction should partially contribute to the overall results. Therefore, we perform the global average pooling operation on $\{\bm{x}_1, \bm{x}_2, ..., \bm{x}_L\}$ to calculate the quiz vector as: 
\begin{equation}  \label{eq4}
	\bm{q} =  \sum_{l=1}^{L}(\bm{x}_l) / L. \\
\end{equation}
where we have $\bm{q} \in \mathbb{R}^{d_q}$ ($d_q$ equals to $d_i$) denotes the quiz vector. Through the above modeling process, $\bm{q}$ contains both the intra-quiz short-term knowledge influence and the student's overall knowledge state on the quiz-related KCs.

\begin{algorithm}[t]  
	\caption{The QKT Model.}  
	\label{a1}  
	\begin{algorithmic}[1]  
		\Require  
		The student's quiz-based interaction sequence,  $U = \{q_1, q_2, ..., q_J\}$;
		The student's interactions in each quiz,  $q = \{i_1, i_2, ..., i_L\}$;
		The exercise $e$ and the student's answer $a$ for each interaction;
		The Q-matrix that indicates the exercise-KC relations.
		\Ensure  
		The student's knowledge state, $\bm{h} \in \mathbb{R}^{d_h}$.
		\State compute the complete exercise vector for all exercises in the student's interactions by Eq. (\ref{eq1}); 
		\State obtain the interaction vector for all the student's interactions from Eq. (\ref{eq2}); 
		\State compute the adjacent gate by Eq. (\ref{eq3}); 
		\State obtain the quiz vector for all the student's finished quizzes from Eq. (\ref{eq4}); 
		\State measure the inter-quiz knowledge substitution by Eq. (\ref{eq5}); 
		\State compute the recency-aware attention value by Eq. (\ref{eq6}),  Eq. (\ref{rencecy}); 
		\State measure the inter-quiz knowledge complementarity by Eq. (\ref{eq7}),  Eq. (\ref{eq8});
		\State integrate the student's knowledge state by Eq. (\ref{eq9}); \\
		\Return $\bm{h}$
	\end{algorithmic}  
\end{algorithm}

\subsection{The Inter-quiz Modeling Module}
In this module, we turn our attention from one quiz to multiple quizzes and try to measure the inter-quiz long-term knowledge integration, i.e.,
integrating different quiz vectors together to represent students' dynamic knowledge states. Specifically, there are two main forms of the inter-quiz long-term knowledge integration: the knowledge substitution and the knowledge complementarity. We will first separately assess them and then combine them together.

\subsubsection{The Knowledge Substitution}
The knowledge substitution means that students' interactions on previous quizzes will be replaced by recent ones, which often appears in quizzes that have the same or similar KCs. For example, a student may get poor performance on the quizzes related to the KC: \textit{Venn Diagrams} when he learned this KC at the beginning. However, after a period of studying, he can achieve perfect performance on subsequent quizzes so that his interactions at the beginning quizzes are unreliable and should be updated by the recent interactions.

To measure the above knowledge substitution, we directly utilize the Gated Recurrent Units (GRU) \citep{cho2014learning} to model students' quiz sequences $\{\bm{q}_1, \bm{q}_2, ..., \bm{q}_J\}$ as follows:
\begin{equation} \label{eq5}
	\begin{aligned}
		\bm{\varGamma}_r &=  \sigma(\bm{W}_3^T(\bm{sub}_{j-1} \oplus \bm{q}_j) + \bm{b}_3), \\
		\bm{\varGamma}_u &=  \sigma(\bm{W}_4^T(\bm{sub}_{j-1} \oplus \bm{q}_j) + \bm{b}_4),  \\
		\bm{\tilde{sub}}_j &= tanh(\bm{\varGamma}_r \cdot \bm{W}_5^T(\bm{sub}_{j-1} \oplus \bm{q}_j) + \bm{b}_5), \\
		\bm{sub}_j &= (1 - \bm{\varGamma}_u) \cdot \bm{sub}_{j-1} + \bm{\varGamma}_u \cdot \bm{\tilde{sub}}_j, \\
	\end{aligned}
\end{equation}
where we have $\bm{sub}_{j-1} \in \mathbb{R}^{d_q}$ denotes the summary of all previous quizzes, $\bm{\varGamma}_r$ is the reset gate that determines how to combine the present quiz vector and previous memories $\bm{sub}_{j-1}$, $\bm{\varGamma}_u$ is the update gate that controls how much previous memories will be preserved. 
$tanh$ is the activation function, $\bm{W}_3 \in  \mathbb{R}^{2d_q \times d_q}$,  $\bm{W}_4 \in  \mathbb{R}^{2d_q \times d_q}$,  $\bm{W}_5 \in  \mathbb{R}^{2d_q \times d_q}$, $\bm{b}_3 \in \mathbb{R}^{d_q}$, $\bm{b}_4 \in \mathbb{R}^{d_q}$, $\bm{b}_5 \in \mathbb{R}^{d_q}$ are trainable parameters. After processing all quizzes in order, we can get the vector $\bm{sub}_J$ that captures the inter-quiz long-term knowledge substitution, which memories more students' interactions on recent quizzes and forgets more on the remote quizzes.  

\subsubsection{The Knowledge Complementarity}
In contrast to the knowledge substitution, knowledge complementarity means that students' interactions on previous quizzes will be effective in parallel with the interactions on the subsequent quizzes. Specifically, in all the quizzes completed by the student, we have measured the knowledge substitution for the quizzes with overlapping KCs. However, there are also many quizzes that focus on different KCs, and students' interactions on these quizzes should complement each other for a more comprehensive knowledge state retrieval.

To measure the above knowledge complementarity, we present the self-attentive encoder to capture the dependency of different quizzes.
Specifically, for each quiz vector $\bm{q}_j$,  we first utilize three embedding layers to respectively project $\bm{q}_j$ into the query vector $\bm{\tilde{q}}_j \in \mathbb{R}^{d_q \times 1}$, the key vector $\bm{\tilde{k}}_j \in \mathbb{R}^{d_q \times 1}$ and the value vector $\bm{\tilde{v}}_j \in \mathbb{R}^{d_q \times 1}$. Then, the dot-product attention value $\alpha_{jj'}$ between $\bm{q}_j$ and $\bm{q}_{j'}$ is calculated as:
\begin{equation} \label{eq6}
	\alpha_{jj'} = \text{softmax}(\bm{\tilde{q}}_j \cdot \bm{\tilde{k}}_{j'}) = \frac{\text{exp}(\bm{\tilde{q}}_j \cdot \bm{\tilde{k}}_{j'})}{\sum_{j' = 1}^J \text{exp}(\bm{\tilde{q}}_j \cdot \bm{\tilde{k}}_{j'})}. \\
\end{equation}

However, $\alpha_{jj'}$ only measure the similarity of $\bm{q}_j$ and $\bm{q}_{j'}$ as their attention weight, without considering the order of a specific quiz in the whole quiz sequence. Actually, learning is temporal and the recent events have more influence on students, which is also known as the recency effect \cite{Glanzer_1966}. Therefore, we argue that it is necessary to weight more recent quiz more heavily and further propose the recency-aware attention mechanism, which adds two recency-aware terms to the attention value $\alpha_{jj'}$ as:
\begin{equation} \label{rencecy}
	\begin{aligned}
		\beta_{j'}^1 &= \text{softmax}(\gamma j') = \frac{\text{exp}(\gamma j')}{\sum_{j' = 1}^J \text{exp}(\gamma j')}, \\
		\beta_{j'}^2 &= \text{softmax}(\gamma (J-j')) = \frac{\text{exp}(\gamma (J-j'))}{\sum_{j' = 1}^J \text{exp}(\gamma (J-j'))}, \\
		\tilde{\alpha}_{jj'} &= \alpha_{jj'} + \beta_{j'}^1 - \beta_{j'}^2, \\
		% &= \alpha_{jj'} + \text{softmax}(\frac{j'}{\gamma}) - \text{softmax}(\frac{J - j'}{\gamma}) \\
		%	&=\alpha_{jj'} + \frac{\text{exp}(\frac{j'}{\gamma})}{\sum_{j' = 1}^J \text{exp}(\frac{j'}{\gamma})} - \frac{\text{exp}(\frac{J - j'}{\gamma})}{\sum_{j' = 1}^J \text{exp}(\frac{J - j'}{\gamma})}, \\
	\end{aligned}
\end{equation}
where $\gamma$ is a constant parameter, which scales the value of the recency-aware terms $\beta_{j'}^1$ and $\beta_{j'}^2$ to match the dot-product attention value $\alpha_{jj'}$. Noting that through adding $\beta_{j'}^1$ and subtracting $\beta_{j'}^2$, we just move part of the dot-product attention value from the front quizzes to the behind quizzes in the student's quiz sequence, realizing the assumption that the more recent quiz matters more. The total value of the dot-product attention is not changed, i.e., $\sum_{j'=1}^{J}\tilde{\alpha}_{jj'} = \sum_{j'=1}^{J}\alpha_{jj'} = 1$.  

Subsequently, $\tilde{\alpha}_{jj'}$ will be multiplied to $\bm{\tilde{v}}_j$ to get the output as a weighted sum of the values: 
\begin{equation}   \label{eq7}
	\bm{z}_{j} = \sum_{j' = 1}^J\tilde{\alpha}_{jj'}\bm{\tilde{v}}_{j'} , \\
\end{equation}
where we have $\bm{z}_{j} \in \mathbb{R}^{d_q \times 1} $ denotes the quiz vector for $q_j$, which includes the knowledge complementarity with other quizzes. Finally, similar to intra-quiz interactions integration in Section \ref{413}, we perform global average pooling on $\{\bm{z}_1, \bm{z}_2, ..., \bm{z}_J\}$ as: 
\begin{equation}    \label{eq8}
	\bm{com}_J =  \sum_{j=1}^{J}(\bm{z}_j) / J. \\
\end{equation}
Then, we can get the vector $\bm{com}_J$ that captures the inter-quiz long-term knowledge complementarity.

\subsubsection{Integrating the knowledge state}
After calculating $\bm{sub}_J$ and $\bm{com}_J$ that respectively model the inter-quiz long-term knowledge substitution and complementarity, we need to further combine them together to output students' knowledge states across different quizzes. In QKT, we utilize a simple but effective way of addition to integrate them:
\begin{equation}    \label{eq9}
	\bm{h} =  \bm{W}_6^T(\bm{sub}_J +\bm{com}_J) + \bm{b}_6, \\
\end{equation}
where we have $\bm{h} \in \mathbb{R}^{d_h}$ denotes the knowledge state vector, $\bm{W}_6 \in  \mathbb{R}^{d_q \times d_h}$ and $\bm{b}_6 \in \mathbb{R}^{d_h}$ are trainable parameters.

\section{Model Learning}

To train all embeddings, weight matrices, and bias terms in QKT, we first use $\bm{h}$ to predict the student's answer on the exercise $e_{n}$ in future quizzes, and then choose the cross-entropy log loss between the predicted answer $y_{n}$ and the student's actual answer $a_{n}$ as the objective function, as follows:

\begin{equation}
	\begin{aligned}
		y_{n} &=  \sigma(\bm{W}_8^T(\bm{W}_7^T(\bm{\tilde{e}}_{n} \cdot \bm{h} \oplus \bm{\tilde{e}}_n \oplus \bm{h}) + \bm{b}_7) + \bm{b}_8), \\
		\mathbb{L}(\theta) &= - \sum(a_{n}\log y_{n} + (1 - a_{n})\log(1 - y_{n})) + \lambda_{\theta}||\theta||^2, \\
	\end{aligned}
\end{equation}
where we use both the multiplication (i.e, $\bm{\tilde{e}}_{n} \cdot \bm{h}$) and the concatenation (i.e,  $\bm{\tilde{e}}_n \oplus \bm{h}$) to integrate the prediction vector,
$\bm{W}_7 \in  \mathbb{R}^{d_h \times d_h}$, $\bm{W}_8 \in  \mathbb{R}^{d_h \times 1}$, $\bm{b}_7 \in \mathbb{R}^{d_h}$ , $\bm{b}_8 \in \mathbb{R}^{1}$ are trainable parameters. $\theta$ denotes all trainable parameters in QKT and $\lambda_{\theta}$ is the regularization c. The objective function will be minimized using Adam optimizer \cite{kingma2014adam} on mini-batches.

\section{Experiments}

%order  num_quiz case
In this section, we first introduce the public real-world datasets utilized in our experiments. Then, we conduct experiments to evaluate the effectiveness of QKT with the aim of answering the following research questions:
\begin{itemize}[leftmargin=*]
	\item{\textbf{RQ1:}} Does our presented QKT model outperform existing methods on the student performance prediction task?
	\item{\textbf{RQ2:}} How do the different components in QKT impact its performance respectively?
	\item{\textbf{RQ3:}} How about the impact of a varying number of quizzes for each student on QKT?
	\item{\textbf{RQ4:}} How about the impact of a varying length of interactions for each quiz on QKT?
\end{itemize}

\subsection{Datasets} \label{datasets}
Three public real-world datasets are used for evaluating in our experiments: (1) Assist2012\footnote{https://sites.google.com/site/assistmentsdata/datasets/2012-13-school-data-with-affect}; (2) Eedi2020\footnote{https://eedi.com/projects/neurips-education-challenge}; (3) CSEDM\footnote{https://sites.google.com/ncsu.edu/csedm-dc-2021/home}. We give the statistics of all datasets in Table \ref{dataset}. The distributions of the quiz length $L$ and the quiz number $J$ in each dataset are also given in Figure \ref{distribution}, which are various across different datasets. The detailed descriptions of all datasets are:

\begin{itemize}[leftmargin=*]
	\item{\textbf{Assist2012}} is collected from 8th-grade students for the school year 2012-2013 in the ASSISTments math tutoring system \citep{feng2009addressing}. Exercises with similar KCs in ASSISTments are organized as the \textit{assignment} (similar to the quiz), and students need to practice on different \textit{assignments} for obtaining the related knowledge. In our experiments, we have filtered the interactions that the exercise's related KCs are missing. 
	\item{\textbf{Eedi2020}} is published in the NeurIPS 2020 Education Challenge, which contains students’ answers to mathematics questions from Eedi, an OIS which millions of students interact with daily around the globe from school year 2018 to 2020. Exercises in Eedi are organized as different quizzes.
	We used the data for task 3\&4 in this challenge. This dataset has hierarchical KCs, we utilize only the KC in the leaf node for each exercise. 
	\item{\textbf{CSEDM}} is published in the 2nd Computer Science Educational Data Mining Challenge, which is collected from a CS1 course in the Spring and Fall 2019 semesters at a public university in the U.S. It contains the code submissions from students for 50 coding problems in 5 different \textit{assignments}.
	
\end{itemize}

\begin{table}[t]
	\caption{Statistics of all datasets. }
	
	\renewcommand\arraystretch{1.2}
	\centering
	\resizebox{\columnwidth}{!}{
		\begin{tabular}{l|rrr}
			\hline
			\multirow{2}*{Statistics}&\multicolumn{3}{c}{Datasets}\\
			\cline{2-4}  
			& Assist2012 &Eedi2020 & CSEDM \\
			\hline
			\# of students & 29,018 & 4,918      & 840\\
			%	\hline
			\# of exercises &53,091  & 948      & 50\\
			%	\hline
			\# of KCs &  265 & 53    & 50 \\
			
			\# of interactions  & 2,711,813 & 1,382,727  & 38,531  \\
			Avg. interactions per student  &  93.45 & 281.16     & 45.87 \\
			Avg. quizzes per student &  19.02 & 20.54    & 4.73   \\
			Avg. interactions per quiz &  4.93  & 13.74     & 9.70 \\
			\hline
			
		\end{tabular}
	}
	
	\label{dataset}
\end{table}

\subsection{Experimental Settings}

For all students' answering records, We first sorted them by the timestamp of answering. Then we utilized their interactions on all previous $J-1$ quizzes to train the model and predict their performance on the exercises in the last quizzes (i.e., the $J$-th quiz). To ensure the reliability of experimental results, we filtered out the students who answered fewer than 2 quizzes. To set up the training process, we randomly initialized all parameters and embeddings in the uniform distribution \cite{glorot2010understanding}. Experiments on all datasets have been 5-fold cross-validated on students\footnote{We will make the code publicly available upon acceptance}.
The initial learning rate was 1e-3 and we set the learning rate decay of 50\% every three epochs to achieve the optimal point. The mini-batch size was 32. The dimensions $d_{e}$, $d_{i}$, $d_q$, and $d_h$ were uniformly set as 128. The constant parameter $\gamma$ utilized to scale the value of the recency-aware terms in Eq. (\ref{rencecy})  was 1e-5.

\begin{figure}[t]

	\centering
	\subfigure[\label{2012_len} The distribution of the quiz length on Assist2012.]{
		\includegraphics[width=0.44\columnwidth]{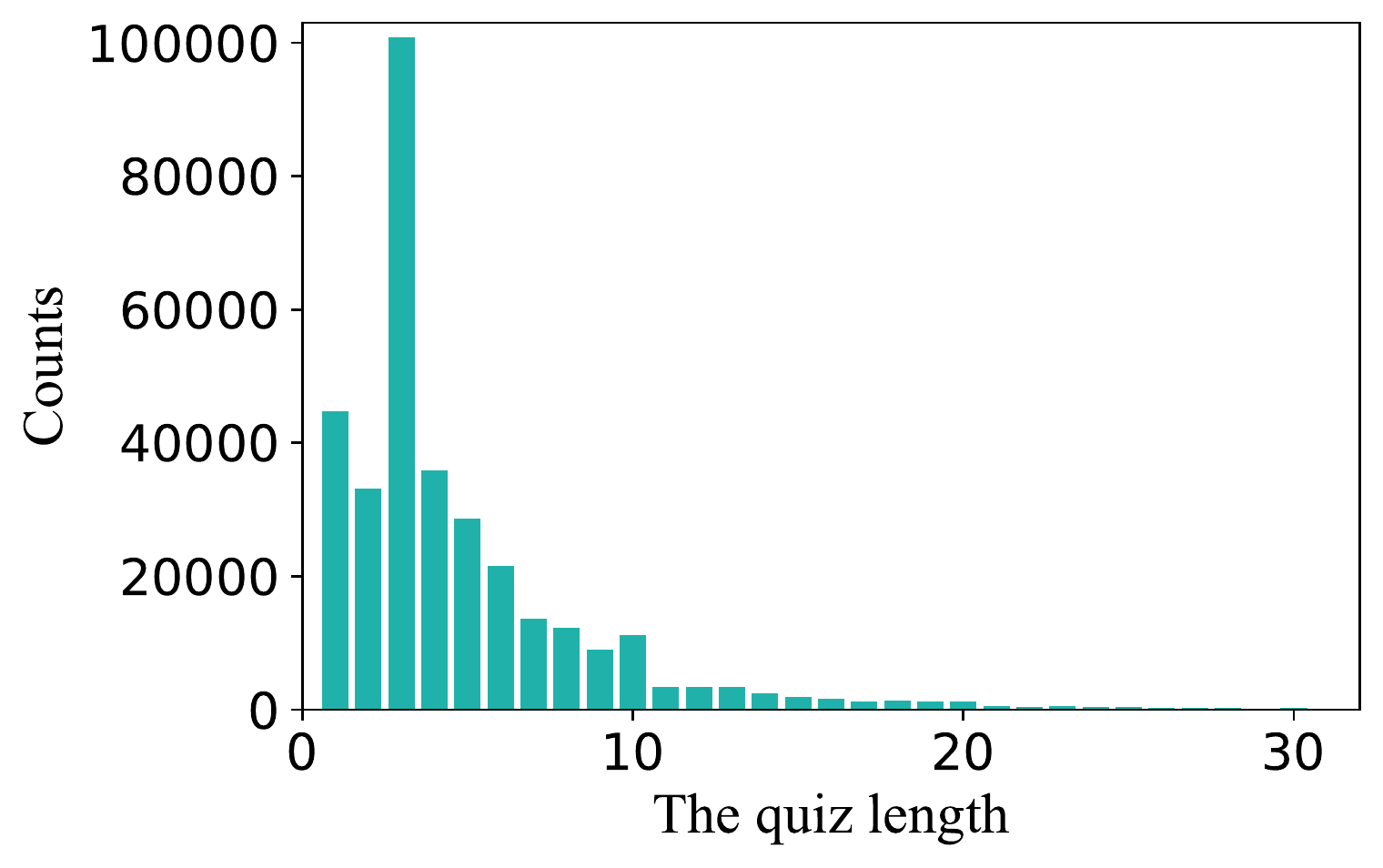}
	}
	\subfigure[\label{2012_num} The distribution of the quiz number on Assist2012.]{
		\includegraphics[width=0.44\columnwidth]{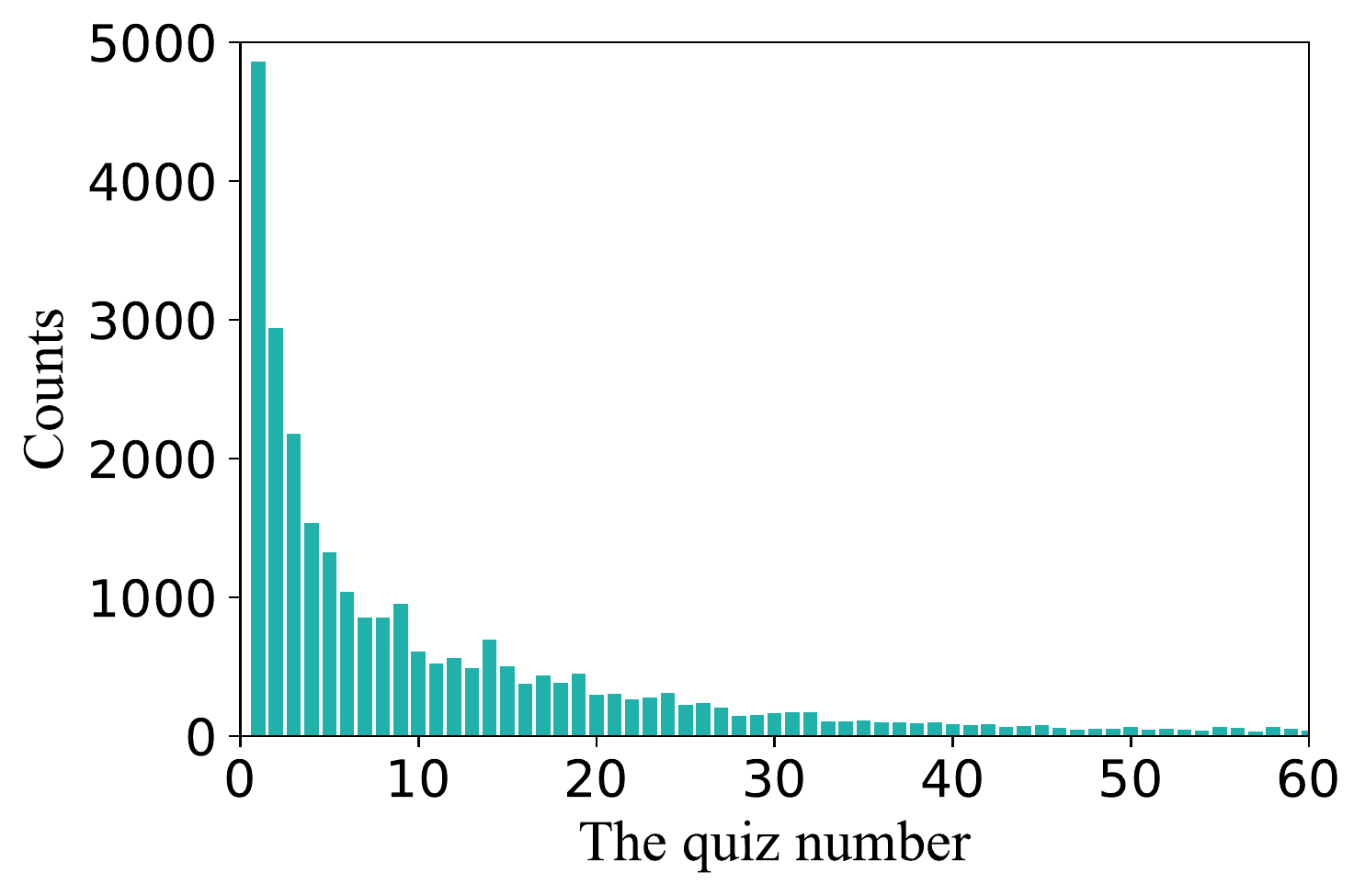}
	}
	\subfigure[\label{eedi_len} The distribution of the quiz length on Eedi2020.]{
		\includegraphics[width=0.44\columnwidth]{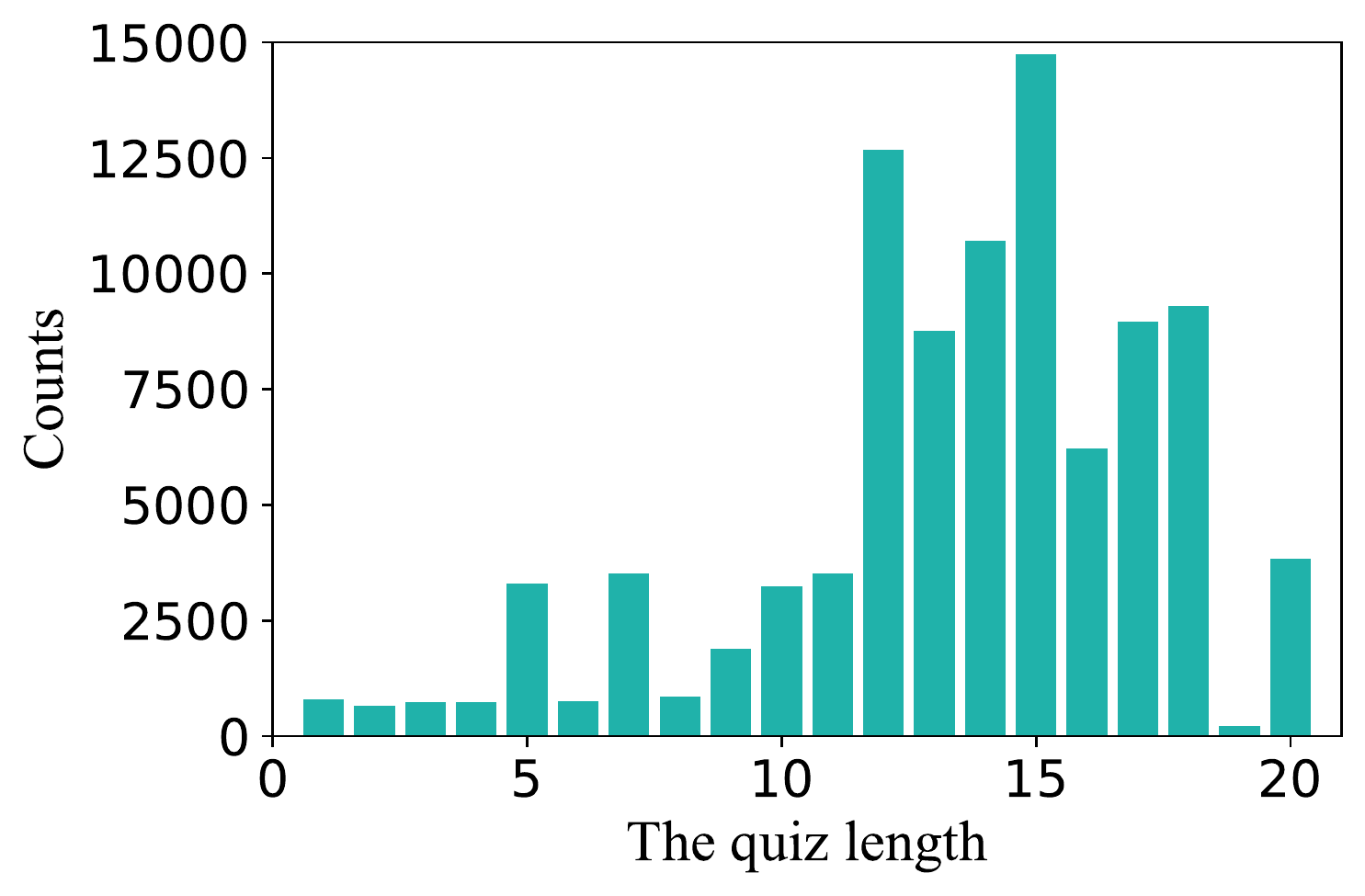}
	}
	\subfigure[\label{eedi_num} The distribution of the quiz number on Eedi2020.]{
		\includegraphics[width=0.44\columnwidth]{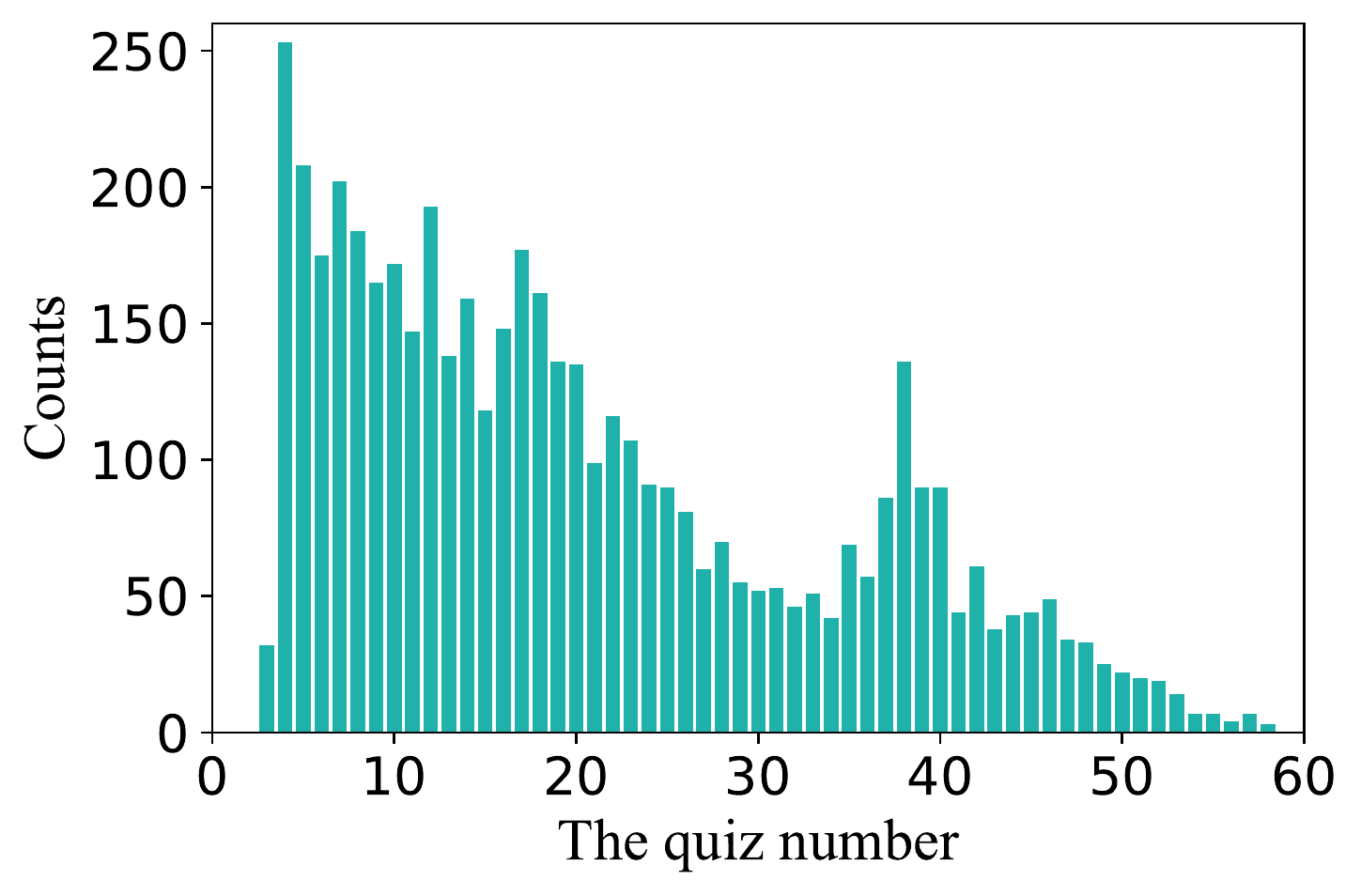}
	}
	\subfigure[\label{cs_len} The distribution of the quiz length on CSEDM.]{
		\includegraphics[width=0.44\columnwidth]{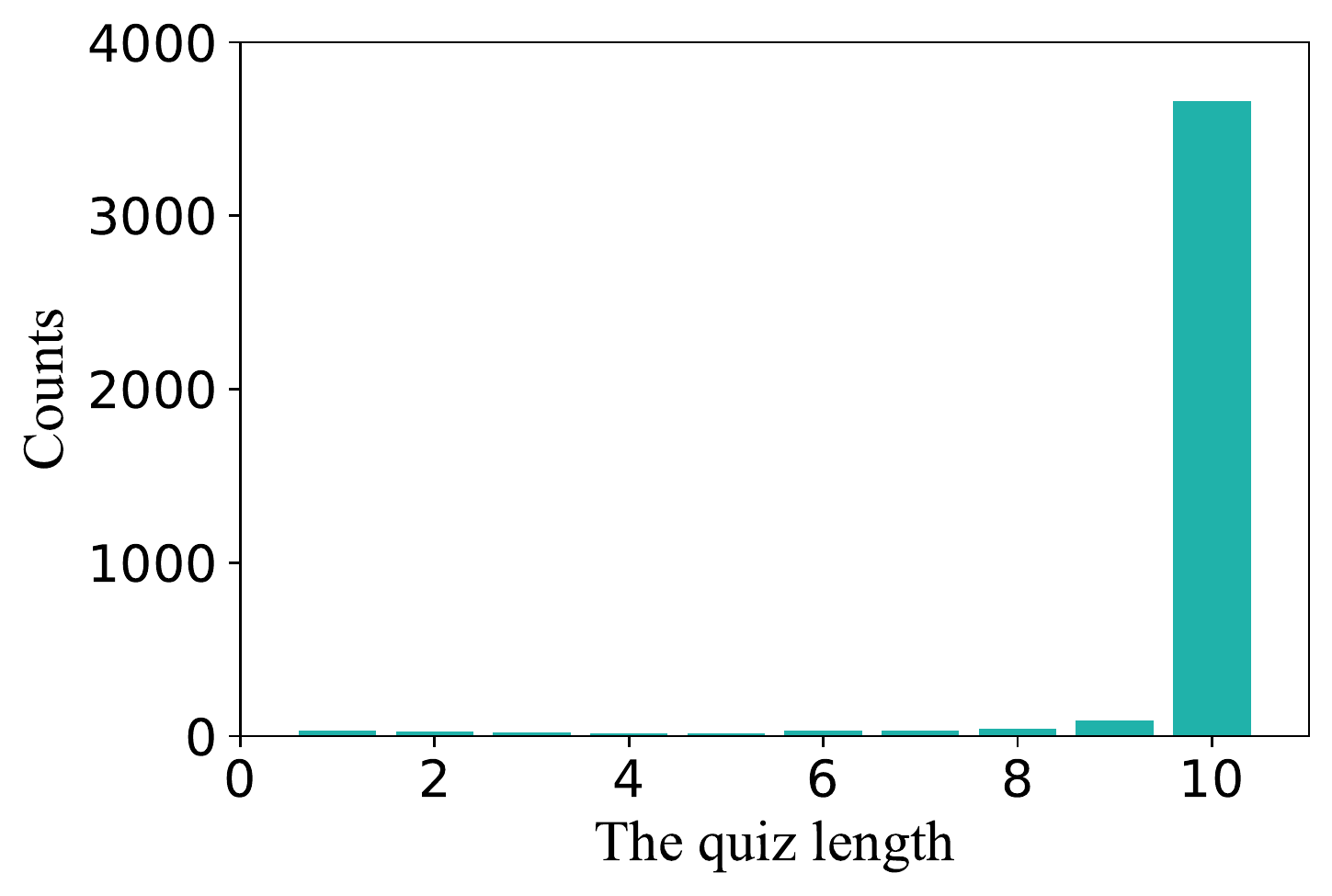}
	}
	\subfigure[\label{cs_num} The distribution of the quiz number on CSEDM.]{
		\includegraphics[width=0.44\columnwidth]{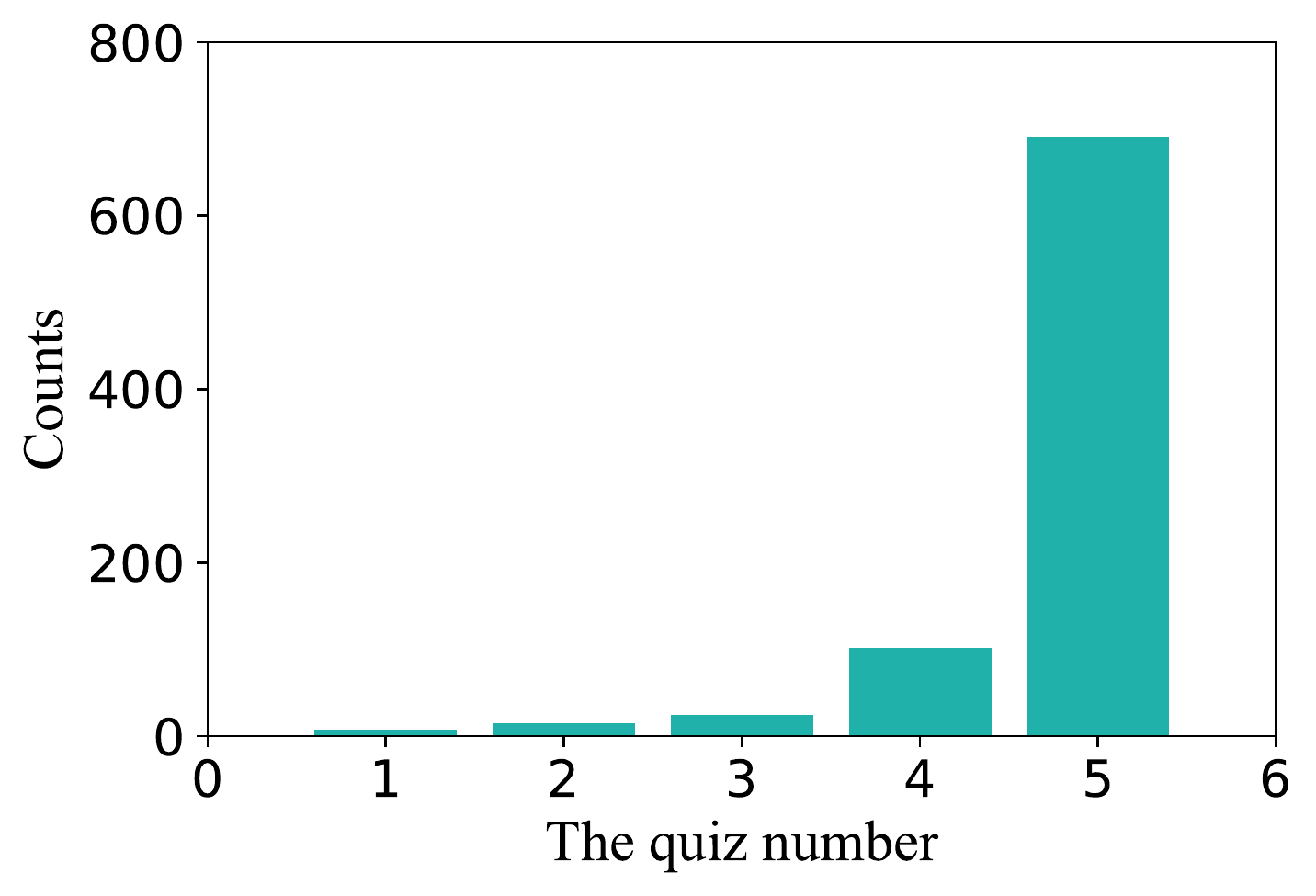}
	}

	\caption{Distributions of the quiz length and the quiz number on all datasets.}
	\label{distribution}

\end{figure}

According to the distributions of the quiz length and the quiz number on all datasets as shown in Figure \ref{distribution}, for Assist2012, the quiz length and the quiz number are both 30 in our experiments. For Eedi2020, we respectively set the quiz length and the quiz number to be 20 and 50.  For CSEDM, the quiz length is 10, and the quiz number is 5.

\begin{table*}[t]
	
	\centering
	\renewcommand{\arraystretch}{1.2}
	\caption{Results ($\times 100$) of all comparison methods on student performance prediction.  We use underline to highlight the best result among all baselines in each column, while bold number shows the best result of the whole column. The error bars after $\pm$ is the standard deviations of 5 evaluation runs for each methods.  The gain shows our model improvement over the best performance of all baselines.}

	\resizebox{\textwidth}{!}{
		\begin{tabular}{l|c|c|c|c|c|c|c|c|c}
			\hline
			\multirow{2}[4]{*}{Methods}   & \multicolumn{3}{c|}{Assist2012} & \multicolumn{3}{c|}{Eedi2020}  & \multicolumn{3}{c}{CSEDM}  \\
			\cline{2-10}        & AUC   & RMSE  & $r^2$    & AUC & RMSE  & $r^2$  & AUC  & RMSE  & $r^2$   \\
			\hline
			IRT   & 70.91$\pm$0.34& 44.98$\pm$0.17 & 12.02$\pm$0.29 & 68.67$\pm$0.67  & 47.22$\pm$0.20 & 10.78$\pm$0.77 & 72.54$\pm$1.67 & 24.40$\pm$0.93 & 2.32$\pm$1.71  \\
			NCD   & 70.66$\pm$0.27& 45.25$\pm$0.30 & 10.99$\pm$0.75  & 69.91$\pm$0.47 & 47.18$\pm$0.23 & 10.94$\pm$0.88 & 74.36$\pm$1.38 & 24.13$\pm$1.05 & 4.16$\pm$0.41  \\
			\hline
			DKT   & 70.28$\pm$0.57& 45.08$\pm$0.28 & 11.63$\pm$0.78 & 73.31$\pm$0.65 & 45.77$\pm$0.26 & 16.20$\pm$0.94 & 83.47$\pm$2.58 & 21.76$\pm$1.70 & 12.44$\pm$6.49 \\
			DKVMN  & 64.36$\pm$0.66& 47.61$\pm$0.67 & 1.45$\pm$2.37 & 73.21$\pm$0.50 & 45.78$\pm$0.23 & 16.14$\pm$0.81 & 83.89$\pm$2.56 & 21.92$\pm$1.61 & 8.82$\pm$6.91  \\
			SAKT   & 66.51$\pm$0.58& 46.57$\pm$0.46 & 5.69$\pm$0.46  & 73.62$\pm$0.29 & 45.52$\pm$0.11 & 17.11$\pm$0.38 & 79.78$\pm$4.33 & 22.31$\pm$1.34 & 8.37$\pm$5.35 \\
			AKT    & 70.99$\pm$0.62& 45.23$\pm$0.59 & 11.03$\pm$2.33  & \underline{74.48$\pm$0.38} & \underline{45.15$\pm$0.15} & \underline{18.43$\pm$0.52} & 84.26$\pm$2.21 & 21.72$\pm$1.38 & 10.40$\pm$6.63 \\
			
			LPKT   & 70.36$\pm$0.76& 44.99$\pm$0.34 & 11.99$\pm$0.94  & 74.38$\pm$0.39 & 45.33$\pm$0.17 & 17.81$\pm$0.60 & 84.96$\pm$2.25 & \underline{21.49$\pm$1.22} & \underline{13.50$\pm$3.43} \\
			DIMKT & \underline{72.04$\pm$0.40}& \underline{44.53$\pm$0.29} & \underline{13.78$\pm$0.79}  & 74.35$\pm$0.49 & 45.35$\pm$0.21 & 17.71$\pm$0.74 & \underline{85.04$\pm$1.74} & 21.50$\pm$1.16 & 13.30$\pm$5.22 \\
			\hline
			QKT  & \textbf{72.71$\pm$0.61} & \textbf{44.28$\pm$0.32} & \textbf{14.72$\pm$0.83} & \textbf{75.23$\pm$0.37} & \textbf{44.83$\pm$0.17} & \textbf{19.68$\pm$0.59} & \textbf{85.40$\pm$1.58} & \textbf{21.39$\pm$1.15} & \textbf{14.31$\pm$3.63} \\
			\hline
			Gain  & 0.93 & 0.56 & 6.82 & 1.00 & 0.71 & 6.78 &  0.42 & 0.47 & 6.00\\
			
			\hline
		\end{tabular}%
	}
	\label{main_exp}%

\end{table*}%

\begin{table*}[t]
	\centering
	\caption{Results ($\times 100$) of ablation experiments on the dataset Assist2012.}

	\renewcommand\arraystretch{1.2}
	%\resizebox{\textwidth}{!}{
		\begin{tabular}{l|cccc|ccc}
			\hline
			Methods   &  \makecell[c]{Intra-quiz\\Influence} & \makecell[c]{Inter-quiz\\Substitution} & \makecell[c]{Inter-quiz\\Complementarity}  & \makecell[c]{Recency-aware \\attention}  & AUC    &  RMSE & $r^2$  \\
			\hline
			%\midrule
			
			QKT w/o KI & \ding{56} & \ding{52}   & \ding{52}   & \ding{52}  & 72.50   & 44.39 & 14.33  \\
			QKT w/o SUB & \ding{52}   &  \ding{56}  & \ding{52}  & \ding{52}  & 71.90   & 44.56 & 13.66 	\\
			QKT w/o COM & \ding{52}  & \ding{52}  & \ding{56}  &  \ding{52}  & 72.49   & 44.37 & 14.40	\\
			QKT w/o RA  & \ding{52}  & \ding{52}  & \ding{52} & \ding{56}  & 72.62   & 44.33 & 14.55	\\
			\hline
			QKT  & \ding{52}  & \ding{52}   & \ding{52}  & \ding{52}  & \textbf{72.71}  & \textbf{44.28} & \textbf{14.72}\\
			\hline
		\end{tabular}%
		%}
	
	\label{ablation}%
\end{table*}

\subsection{Comparison Baselines}
To verify the effectiveness of QKT, we compare it with existing KT methods. All comparison methods are tuned to have the best performances for a fair comparison. All models are completed by Tensorflow and trained on a cluster of Linux servers with the NVIDIA Tesla V100 GPU. The simple introduction of KT baselines are: 
\begin{itemize}[leftmargin=*]
	\item{\textbf{DKT}}
	introduces RNNs/LSTMs to model students' knowledge states in a sequential manner \cite{piech2015deep}.
	\item{\textbf{DKVMN}}
	uses the memory network to store and update the knowledge state  \cite{zhang2017dynamic}. 
	\item{\textbf{SAKT}}
	utilizes the self-attention mechanism to capture the knowledge dependency between student-exercise interactions \cite{pandey2019self}.
	\item{\textbf{AKT}}
	learns context-aware interaction representations by the self-attentive encoder, and measures students' knowledge acquired in the past relevant to the current exercise \cite{CAKT}.
	\item{\textbf{LPKT}}
	models the learning process and calculates students' learning gains and forgetting to assess their knowledge states \cite{shen2021learning}.
	\item{\textbf{DIMKT}}
	measures the influence of the exercise difficulty on the knowledge state and learning process \cite{shen2022assessing}.
\end{itemize}

Moreover, we also compare QKT with two baselines in the area of cognitive diagnosis:
\begin{itemize}[leftmargin=*]
	\item{\textbf{IRT}}
	uses the logistic function to model  students' knowledge states as a continuous variable \cite{hambleton1991fundamentals}. 
	\item{\textbf{NCD}}
	presents neural networks to learn the complex student-exercise interactions \citep{Wang2020neural}.
\end{itemize}

\subsection{Evaluation Metrics}
To evaluate the performance of QKT and all baselines, we use multiple metrics from both regression and classification perspectives. Specifically, from the perspective of a classification task, we utilize Area Under roc Curve (AUC) to measure the effectiveness, and the larger values are, the better the result.
Then, as a regression task, we quantify the distance between the predicted
and actual answers with Root Mean Square Error (RMSE) and the square of Pearson correlation ($r^2$). For the RMSE, smaller values mean better results. The $r^2$ is the opposite, where larger values are better results.

\subsection{Student Performance Prediction (RQ1)}
The student performance prediction task is crucial for evaluating the quality of the captured knowledge state, i.e., correct predictions stand for better estimation of the knowledge states. To evaluate the effectiveness of QKT, we compare it with all baselines on this task. Table \ref{main_exp} gives the experimental results of our model and all baselines, where we can find several significant observations. 

First, QKT outperforms all existing methods on all datasets and all metrics, which indicates that our proposed QKT is effective to capture the quiz-based organization style of students' learning interactions. Second, in contrast to some baselines (such as LPKT) that introduced the answer time and interval time as additional features to model the whole learning process, QKT achieves better performance without using any time information, which further demonstrates the significance and value of exploring students' interaction sequences from the quiz perspective. Noting that the time information is also important for QKT, as this paper mainly focuses on finding and evaluating the necessity of conducting quiz-based knowledge tracing , we leave the extension to time information of QKT as a future work. Third, the performance gains of QKT against the best baseline are positively related to students' average quiz number, which is in line with our intuition as QKT benefits more from the inter-quiz modeling module when there are more quizzes (we will further verify it in section \ref{quiz_number}). For example, the average number of quizzes taken by students in the datasets Assist2012 and Eedi2020 is close (i.e., 19.02 for Assist2012 and 20.54 for Eedi2020, as shown in Table \ref{dataset}), the performance gains of QKT against the best baseline are also close in both datasets. However, students in CSEDM finished less quizzes, and QKT's performance gains are relatively smaller.

\begin{figure*}[t]

	\centering{\includegraphics[width=0.8\textwidth]{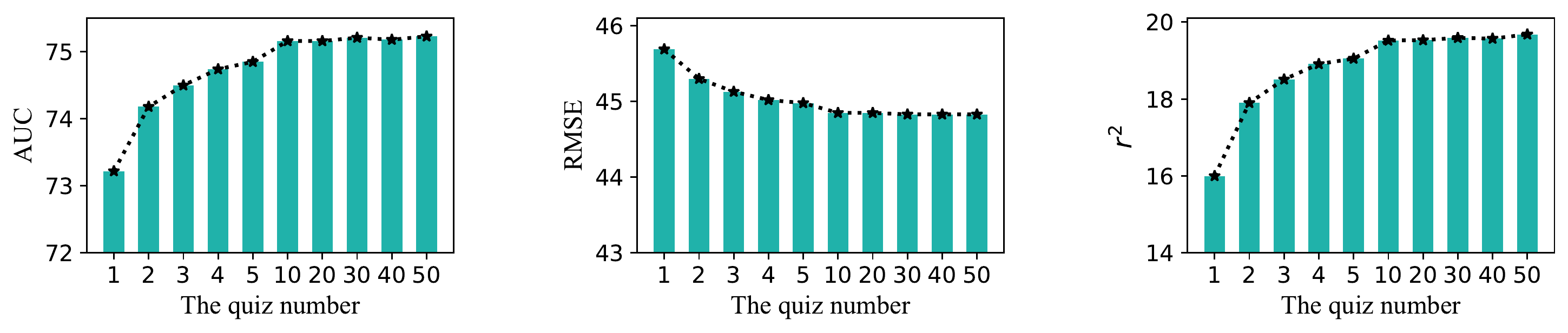}}

	\caption{The performance ($\times 100$) of QKT with different quiz numbers on Eedi2020. }
	\label{num_quiz}

\end{figure*}
\begin{figure*}[t]

	\centering{\includegraphics[width=0.8\textwidth]{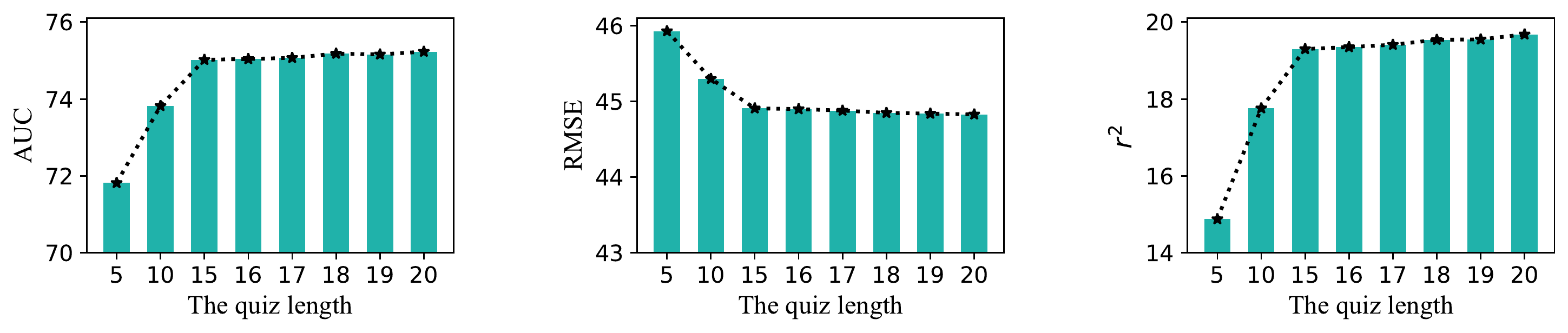}}

	\caption{The performance ($\times 100$) of QKT with different quiz lengths on Eedi2020.}
	\label{len_quiz}

\end{figure*}
\subsection{Ablation Study (RQ2)}
In this section, we conduct ablation experiments to show how different components in QKT affects its performance. The experimental results on Assist2012 are shown in Table \ref{ablation}. We have designed four variations of QKT, each of which takes one component from QKT and reserves others. The details of these variations are:
\begin{itemize}[leftmargin=*,itemsep=2.5pt]
	\item{\textbf{QKT w/o KI}}
	refers to QKT without the knowledge influence modeling, i.e., we removed the adjacent gate and its corresponding operations in QKT w/o KI.
	\item{\textbf{QKT w/o SUB}}
	refers to QKT without the knowledge substitution modeling, where we removed the GRU module and assumed that there was only inter-quiz long-term knowledge complementarity, i.e., students' knowledge state is totally represented by $\bm{com}_j$ in QKT w/o SUB.
	\item{\textbf{QKT w/o COM}}
	refers to QKT without the knowledge complementarity modeling, where we removed the self-attentive encoder and assumed that there was only inter-quiz long-term knowledge substitution, i.e., students' knowledge state is totally represented by $\bm{sub}_j$ in QKT w/o SUB.
	\item{\textbf{QKT w/o RA}} 
	refers to QKT without using the recency-aware attention mechanism, i.e., we removed the recency-aware terms $\beta_{j'}^1$ and $\beta_{j'}^2$ in Eq. (\ref{rencecy}).
\end{itemize}

We can find some interesting conclusions from the results shown in Table \ref{ablation}. First, removing the knowledge substitution modeling leads to the most significant performance decline of QKT, suggesting that equally considering students' previous and recent quizzes with the same or similar KCs heavily damages the knowledge state modeling, as students' interactions at the previous quizzes were not reliable anymore after answering the follow-by quizzes. Second, the inter-quiz knowledge complementarity is also important. If we only model the inter-quiz knowledge substitution, the information contained in quizzes without overlapped KCs will be lost, and the performance of QKT also drops as expected. 
Third, measuring the adjacent influence between interactions within a quiz is necessary, which is critical for intra-quiz short-term knowledge influence modeling. 
Finally, the proposed recency-aware attention mechanism helps to better measure the knowledge substitution, which verifies our assumption that more recent quizzes matter more.

\subsection{The Impact of the Quiz Number (RQ3)} \label{quiz_number}
As this paper focuses on the quiz-based KT, we will further evaluate how different numbers of quizzes for each student (i.e., the quiz number) will affect QKT's performance in this section.
Specifically, we compared the performance of QKT under 10 different quiz numbers on Eedi2020, i.e., 1, 2, 3, 4, 5, 10, 20, 30, 40, 50. The corresponding results are reported in Figure \ref{num_quiz}, where we can directly observe a positive relationship between QKT's performance and the quiz number when the quiz number is small. Moreover, we can get more interesting findings after a detailed look at the figure. First, the performance of QKT tends to be stable when the quiz number reaches a threshold, which is approximately 10 in Figure \ref{num_quiz}. This phenomenon reflects certain marginal effects of the quiz number, which inspires us to design more effective quiz combinations rather than asking students to finish more quizzes for better estimating their knowledge states.
Second, looking at the leftmost bar in in each subplot of Figure \ref{num_quiz}, QKT gets poor performance if we model only one quiz. However, there is a huge promotion if we consider more than one quizzes. This phenomenon suggests that QKT benefits greatly from modeling the inter-quiz long-term knowledge integration. 

\subsection{The Impact of the Quiz Length (RQ4)}
In contrast to the quiz number, we also evaluate QKT's different performance of QKT under various quiz lengths (i.e., the length of interactions in each quiz). Specifically, we compared the performance of QKT under 8 different quiz lengths on Eedi2020, i.e., 5, 10, 15, 16, 17, 18, 19, 20. The corresponding results are reported in Figure \ref{len_quiz}, where we can find more interactions in a quiz bring better performance as expected. The reason is that more interactions contain more reliable information about students' knowledge states of the quiz-related KCs. Similar to the quiz number, we can find certain marginal effects of the quiz length, i.e., QKT's performance grows very slowly after the quiz length reaches a threshold (about 15 in Figure \ref{len_quiz}), which can instruct us to design more effective quizzes with less exercises to improve students' learning efficiency.

\section{Conclusions and Future Works}
In this paper, we focused on students' interaction sequences from the quiz-based perspective for the KT task, and proposed a novel Quiz-based Knowledge Tracing (QKT) to model the interaction sequence in the quiz-based organization. We first analyzed and summarized the feature of students' quiz-based interaction sequences, i.e., it was continuous over a short period of time within a quiz and discrete with certain intervals across different quizzes.  Then, on the basis of above features, we respectively considered the intra-quiz short-term knowledge influence and inter-quiz long-term knowledge substitution and complementarity, and further designed corresponding module in QKT to measure and integrate them to monitor students' dynamic knowledge states. Finally, we conducted extensive experiments on three public real-world datasets to evaluate the effectiveness of QKT, which indicated that QKT achieved better performance than existing best methods. Further analyses on the effects of the quiz number and the quiz length demonstrated that QKT had the potential to benefit the quiz design. 
In the future,  we will further explore utilizing the time information (such as answer time and interval time) for more precise intra- and inter-quiz relation modeling. Besides,  we will attempt to introduce pre-defined knowledge relation of different quizzes to explicitly measure the inter-quiz knowledge integration.

\bibliographystyle{unsrtnat.bst}
\bibliography{cite_file}

\end{document}